# Brain Tumor Segmentation: A Comparative Analysis

**Muhammad Ali Qadar[1], Yan Zhaowen[2]**

**[1] School of Electronics and Information Engineering, Beihang University
Beijing, 100191, China**

**[2] School of Electronics and Information Engineering, Beihang University
Beijing, 100191, China**

**Abstract**

Five different threshold segmentation based approaches have been reviewed and compared over here to extract the tumor from set of brain images. This research focuses on the analysis of image segmentation methods, a comparison of five semi-automated methods have been undertaken for evaluating their relative performance in the segmentation of tumor. Consequently, results are compared on the basis of quantitative and qualitative analysis of respective methods. The purpose of this study was to analytically identify the methods, most suitable for application for a particular genre of problems. The results show that of the region growing segmentation performed better than rest in most cases.

## 1. Introduction

Image segmentation is the fundamental step in medical image analysis. Segmentation is a procedure to separate similar portions of images showing resemblance in different features like shape, size, color, etc. [1]. For the segmentation of medical images, mostly grayscale images are used.

Tumors are commonly stated as the abnormal growth of tissues [2] and the brain tumor is a diseased part in the body tissues that is an abnormal mass in which growth rate of cells is irrepressible [3]. Due to brain tumor's mortality rate have raised over the past years among young people, therefore this area have gained the attention of researchers. Commonly a tumor could be benign or malignant. Benign tumors are those tumors that remain within the boundaries of the brain, whereas the malignant tumors could extend beyond the brain and affect other parts of the body. These kind of tumors may not be treated because of their aggressive nature. Now a days imaging is playing a vital role in diagnosis of the brain tumor in early stages before they become intractable, thus saving many lives. Different techniques have been developed to detect the tumors, like CT, MRI, EEG (electroencephalography) etc. The MR imaging method is the best due to its higher resolution and enhanced quality [4]. Automatic detection of brain tumor is a challenging task because it involves pathology, functional physics of MRI along with intensity and shape analysis of MR image, because Tumor shape, size, location and intensity varies for each infected case [5].

Image segmentation algorithms are based on gray-level values of the pixels, sudden changes in the gray-level and similarity between pixels regions are the basis for segmentation of an image [6]. Many different methods have been proposed for the segmentation of brain tumor from MR images, a bounding box method using symmetry presented by Baidya Nath Saha et.al [7] to segment out tumors from brain MR images, Knowledge based techniques presented by Matthew C. Clark et.al [8] describe and compare results based on supervised and unsupervised clustering. C.L. Biji et.al [9] proposed fuzzy thresholding technique for brain tumor segmentation. Jianping Fan et.al [10] proposed a seeded region growing method in which seed selection and pixel labeling problem are addressed. Yu-len huang and Dar-ren chen [11] proposed segmentation based on a Watershed method for identifying the breast tumors. Nelly Gordillo et.al. [12], presented a review of the most relevant brain tumor segmentation methods, the paper successfully highlights the systematic evidences of the usefulness and limitations of threshold based, region based, pixel based and model based semi-automated and fully automated segmentation techniques. According to [12] in medical image, semi-automated and fully automated segmentation methods have gained the importance due to accuracy in identification, but it's a fact that the end systems are used by the physicians therefore there is a surprising lack of compatibility between large computer vision based frameworks and the low-level methods employed for segmentation. The other reason is that these approaches are still not capable to gain acceptance among pathologist for everyday clinical tasks due not having any standardized procedures. Therefore, these approaches need to be compared with real world medical issues to address problems of segmentation with best suitable approaches.

Image dataset employed here could operate on the gray-level image segmentation algorithms. In this paper, a

This work was supported by the National Natural Science Foundation of China (NSFC) under Grant 61271044.

comparison between seeded region growing, global thresholding, histogram thresholding, fuzzy c-mean, and watershed based brain tumor segmentation methods are taken into account. Statistical and visual analysis is performed to figure out the best method. This research could help clinicians in surgical planning, treatment planning, and accurately segmenting the tumor part with the most accurate method.

The outline of the paper is as follows. Section II explains existing segmentation techniques. Section III presents a performance analysis and experimental results performed during segmentation. Finally, section IV concludes the paper.

## 2. Segmentation Schemes

A set of 40 images have been taken from self-collected dataset, tumor region is extracted in all of these images using segmentation methods, Figure.1. Shows scheme of segmentation criteria is given in the flow graph below. A flowchart of the adopted scheme is presented below accompanied by a brief overview of segmentation methods

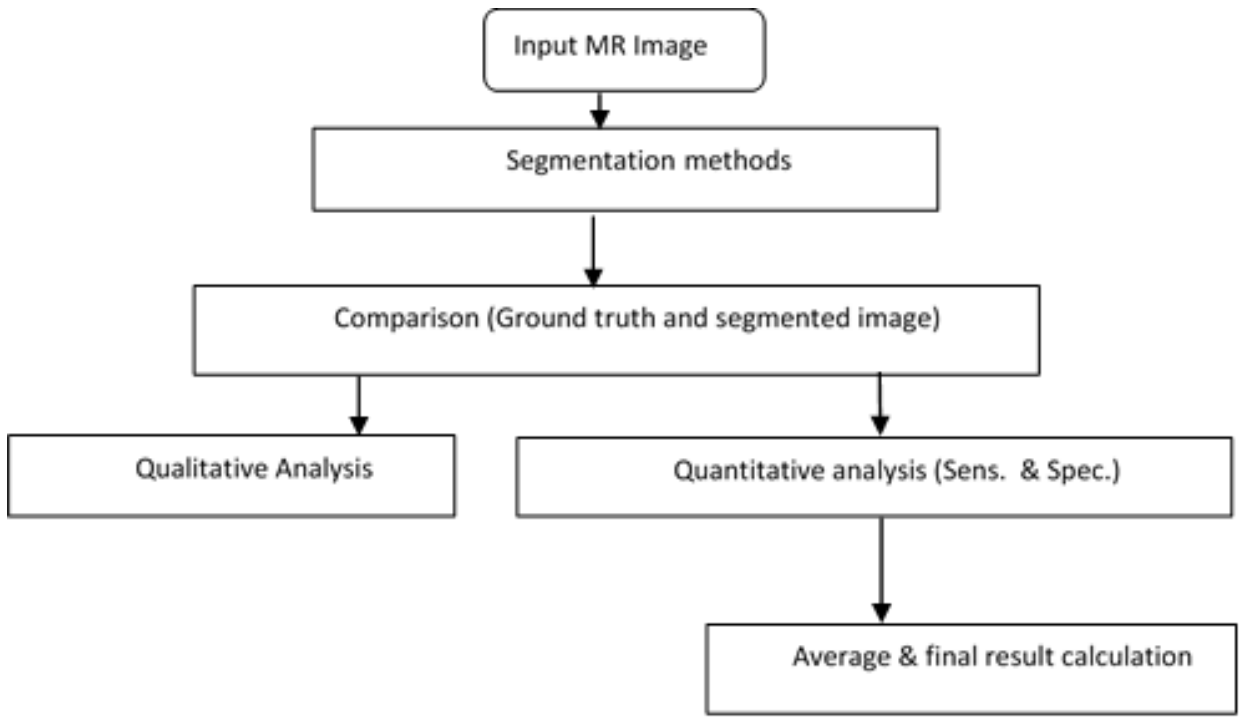

Figure 1: Comparison Scheme for Segmentation

### 2.1 Seeded Region Growing

Segmentation carried out based on set of point known as seeds, the grouping of pixels into regions based on seed points in which region grow by appending seeds to the neighboring pixels [12][1][13]. For the accurate segmentation of regions each connected component of the region should meet exactly one seed [10]. This process of region growing would not stop till all the pixels are grouped into regions by comparing seed pixel with all neighboring pixels [14]. The major issue encountered, is the selection of seed point that is selected manually or by automatic seed selection criteria, also region growing involves high-level of knowledge for semantic image segmentation to explore the seed selection to get more accurate segmentation of regions [10]. For the interpretation, image should be partitioned into meaningful regions which are related to objects in the targeted image. Pixels corresponding to object in image are grouped together and marked. If there are number of seeds grouped into n regions, $R_1, R_2, R_3 \ldots\ldots R_i$ during each iteration there is an addition of one pixel into these regions. Now consider the state of $R_i$ after m steps and $L$ is the set of unallocated pixels

$$L = \left\{ x \notin \bigcup_{i=1}^{n} R_i \mid \mathrm{I}(x) \cap \bigcup_{i=1}^{n} R_i \neq \emptyset \right\} \qquad (1)$$

Where $I(x)$ is the contiguous neighbor of pixel $x, x \in L$ means that $I(x)$ maps exactly one $R_i$ whereas $i(x) = \{1, 2 \ldots. n\}$ with $\mathrm{I}(x) \cap R_i \neq \emptyset$ and $\delta(x)$ is a measure of change in $x$ from the region next to it. $\delta(x)$ could be defined as

$$\delta(x) = \left| g(x) - \mathrm{mean}_{y \in R_{i(x)}}[g(y)] \right| \qquad (2)$$

Where $g(x)$ is the gray-scale value of image point $x$. Here $x$ is the specific pixel that append to existing boundary pixels. Also $x \in L$ follows that

$$\delta(x) = min_{x \in L}\{\, \delta(x)\} \qquad (3)$$

and attach $x$ to $R_l$, this procedure goes on till all pixels are assigned. The above equations (1) and (2) guarantees the regions that are segmented out would be as similar given the connectivity limitations [12].

### 2.2 Threshold based Segmentation

This is simplest image segmentation technique for partitioning images directly into regions based on intensity values with one or more thresholds [1] [12]. Thresholding could be categorized into global or local thresholding based on the number of thresholds selected. Segmentation of images having more than two kind of regions corresponding to different objects regarded as local thresholding [12]. Based on intensity of image, light objects in the dark background are segmented out by selecting specific threshold value $TH$, those pixels that are above threshold are treated as 1 and those are below threshold are set to zero in image $f(x, y)$ with $g(x, y)$ as the segmented image

$$g(x, y) = \begin{cases} 1 & if\ f(x,y) \geq TH \\ 0 & otherwise \end{cases} \qquad (4)$$

Pixels with value 1 corresponds to region of interest (ROI) whereas remaining pixels that are set to zero corresponds to background of image. This type of thresholding technique is known as global thresholding. Brain tumor segmentation using thresholding is carried out to extract the tumor accompanied by fine tuning the segmentation processing employing morphological operations [5]. As this method is based on thresholding therefore results are not much accurate in segmentation, so to enhance the accuracy of this method after the post-processing step region growing is applied to get more accurate extraction of area of tumor.

## 2.3 Watershed Segmentation

Watershed is a geological term described as a narrow hilly area that partitioned two bodies of water, and the area draining into bodies of water or rivers are known as catchment basin [1].

Grouping of pixels based on their intensities is another definition of watershed segmentation [15]. Its gradient-based segmentation technique, gradients are heights in which water rise until local maxima and two bodies of water form a dam. The image is segmented by the dams are 'watersheds' and segmented regions are known as catchments [16]. Elevation to corresponding position is represented by intensity value pixel. Watershed lines are determined on topographic surface by watershed transformation [11]. This algorithm after performing threshold segmentation perform watershed segmentation to mark the tumor region of brain. Then morphological operations helps to detect the final region of tumor. As concerned with preprocessing step the original tumor image is converted from RGB to gray-scale image, then it is filtered through a high pass filter of mask

$$\begin{bmatrix} -1 & 2 & -1 \\ 0 & 0 & 0 \\ 1 & -2 & 1 \end{bmatrix} \quad (5)$$

Which is then passed through the median filter to remove unwanted noise. After thresholding the image, watershed segmentation is applied to detect the boundary regions of the tumor. The post processing of segmented image with following basic functions, dilation and erosion is as follows

$$A \oplus B = \{z|(\hat{B})_z \cap A \neq \phi\} \quad (6)$$

Where $\phi$ the empty set and B is the structuring element. Erosion is given as follows

$$A \ominus B = \{z|(B)_z \cap A^c = \phi\} \quad (7)$$

Both the operations helps to remove the useless information from the segmented image and resulting in final detected tumor image. Watershed segmentation is a powerful tool for image segmentation, it provides closed contours as well as require less computation time but one disadvantage of this technique is over-segmentation, for that marker-based watershed segmentation technique is presented in [17].

## 2.4 Fuzzy C-Mean

The Roman Fuzzy c-Mean is a popular technique for brain tumor segmentation in area of unsupervised image segmentation [12]. FCM provides high degree of membership to every pixel that are close to the threshold point. As in Otsu [18] method output relies on input so a proper initial threshold is required [18] [9]. Initial threshold could be calculated using following formula

$$T = max(grayValue)/2 \quad (8)$$

After selection of initial threshold Fuzzy c-Mean thresholding [9] is performed to extract the tumor region and then post processing using morphological operations are applied. This algorithm takes the gray values as feature, the objective function that is need to be minimized is given as follows.

$$J = \sum_{i=1}^{2} \sum_{j=0}^{L-1} h_j \, \mu_i(j)^\tau d(j, v_i)^2 \quad (9)$$

In (9) objective function is minimized performing iteratively means by (10) and then get the membership equation (11)

$$v_i = \frac{\sum_{j=1}^{L} j h_j \mu_i{}^\tau(j)}{\sum_{j=1}^{L} h_j \mu_i{}^\tau(j)} \quad (10)$$

$$\mu_o - (j) = \frac{1}{1+\left[\frac{d(j,v_2)}{d(j,v_1)}\right]^{\frac{2}{\tau-1}}} \quad (11)$$

This techniques is found very useful in case where there are large number of objects clustered in image. As the algorithm is robust and complex than simple threshold algorithms therefore it performs better for segmentation of intense part in brain images. The major reason of choosing this algorithm is that it divides the brain structure into types of tissue sets, and degree of belongings is assigned to pixels constrained in specific tissue sets having similarity between regions containing fuzzy membership functions in range 0 and 1. If the value is close to 1 and accurate estimation of cluster centers, the algorithm could converge faster and clustering results comes out better [12].

## 2.5 Histogram Thresholding

This algorithm based on the (i) symmetrical structure of the brain, (ii) pixel intensity of image and (iii) binary image conversion [4]. Image is partitioned into two halves, and histogram of each half is computed for comparison of the two histograms. The threshold based on comparison of two histograms is selected and targeted brain image is segmented based on the computed threshold, Finally Crop the image and calculate the area of tumor [4]. Calculation of area of tumor is carried out by calculating the pixel per inch of segmented image that could be calculated by following equation

$$P = \frac{1}{(Horizontal\ dim)*(Vertical\ dim)} \quad (12)$$

And area is calculated by following equation

$$Tumor\ Area = P * T \quad (13)$$

Where T is total number of white pixels in the segmented image, where Tumor Area represents the value of tumor in unit length. The speciality of this algorithm is that it deals with symmetrical structure of brain and in very simple way it segments out the brain tumor from the left or right part of image. After the division of brain into two halves it could be concluded, which part of the brain is having more number of pixels with high intensity value [4]. Histogram comparison is process of comparing each bin of one histogram to other and resultant histogram is used to find the Otsu's threshold level [18] to perform thresholding operation as stated in (Equation. 4).

Fast segmentation concluded than previously stated algorithms but degree of accuracy of segmentation is less because when thresholding based methods are used, the user is supposed to troubleshoot the threshold selection sometimes it loose information and sometimes it will include extra pixels in the background that are undesirable. The time computational complexity of each algorithm is given in following figure.2, looking into graph shows region growing have the highest complexity of processing than other algorithms

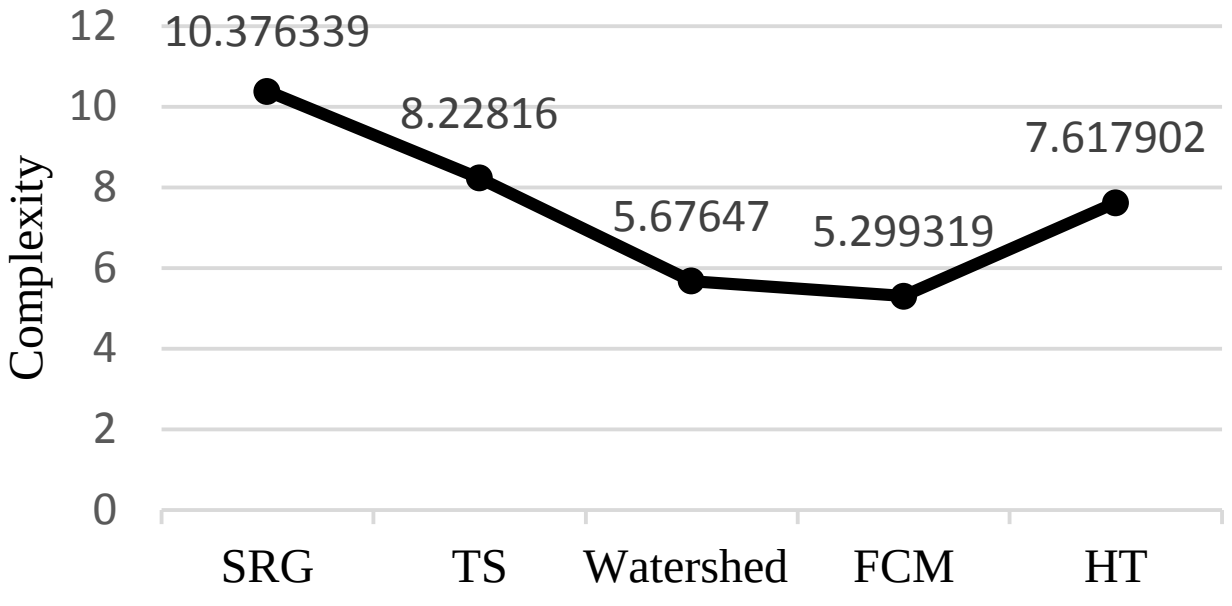

Figure 2: Comparison of Algorithm Complexity

## 3. Experimental Results and Analysis

For the performance analysis a sequence of 40 images have been taken in same class of tumor (benign) segmented with a common threshold value in comparison to radiologist segmented image. For quantitative analysis, number of false negative and false positive are calculated based on number of pixels of interested region (ROI). Four parameter true positive (TP), false positive (FP), true negative (TN), false negative (FN) are calculated by the logical AND between ground truth and segmented image. Sensitivity and specificity in terms of brain tumor region could be defined as, sensitivity is the percentage of patients correctly detected with tumor, whereas specificity is the percentage of patients could not correctly identified with tumor, and F-score measure accuracy of test, it has been reported that higher the value of F-score more accurate is the test [19], formulation as follows:

$$\text{Senstivity} = \frac{\text{TP}}{\text{TP+FN}} \tag{14}$$

$$\text{Specificity} = \frac{\text{TN}}{\text{TN+FP}} \tag{15}$$

$$Precision = \frac{\text{TP}}{\text{TP+FP}} \tag{16}$$

$$\text{F} - \text{score} = \frac{2*(\text{sentivity}*\text{precision})}{\text{sentivity+precision}} \tag{17}$$

Figure.6. illustrates the performance of algorithms in terms of above test parameters. From the performance overview seeded region growing could be predicted as the best one because of its property to segment the homogenous regions, as concerned to nature of medical image dataset, this algorithm ensure its performance in computation time and average accuracy of segmented area of tumor with highest sensitivity having satisfied specificity as shown in figure.6. To address the issue of multiple disjoint tumor parts, a small experiment result is added by setting more than one threshold at the end of figure.8. To compare the disjoint tumor parts, (equation.12) and (equation.13) could be employed to get the non-zero area in segmented image. Figure.4 shows the area calculated for multiple disjoint tumor parts.

All dataset is self-made, it is collected from internet resources, some images are arranged from hospital and all dataset is tested on MATLAB 2012a on core i7 machine. All methods are compared based on the segmentation accuracy of area of tumor [20], comparison of results performed by considering the statistical values and visual comparison of brain tumor images.

Five different threshold based segmentation methods have been implemented in context to the application of detection of region of interest (ROI) from MRI dataset of brain tumor. Contribution in this area of research is done by analysis of image segmentation methods, the statistical, visual, and experimental evidences are provided by comparing segmentation results with ground truth, expert radiologists manually segmented area of tumor. Visual and statistics results not only simulates the methods in literature but also validates with experimental data that is collected from radiologist. Although all of these methods are compared separately, however an effort of comparison for all these methods is missing in the literature, which.are included in this research. Figure.3. shows statistical percentage values for visual comparison of segmentation done for the primary study of tumor. Figure.4. shows the comparison done based on area of tumor, the accuracy of segmentation method is measured based on the comparison with ground truth image that is segmented manually. Area of tumor is considered as a measure for the performance of the segmentation algorithm, because physicians manually segment the tumor taking area of tumor as a parameter of measurement and algorithm could be compared with segmented value by physician [6][20].

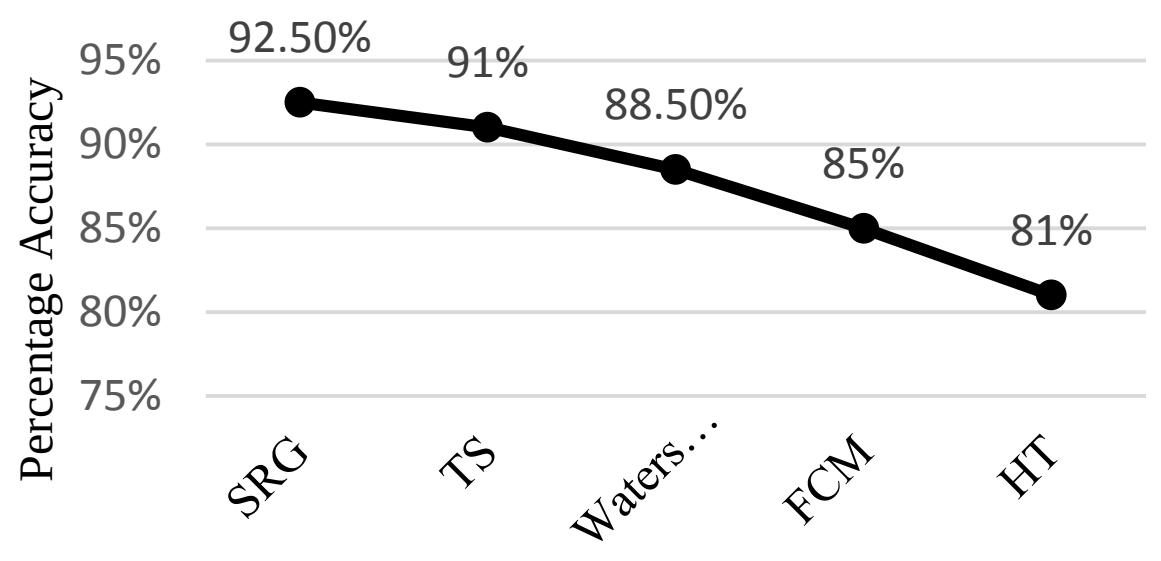

Figure 3: Comparison of Percentage accuracy of Segmentation

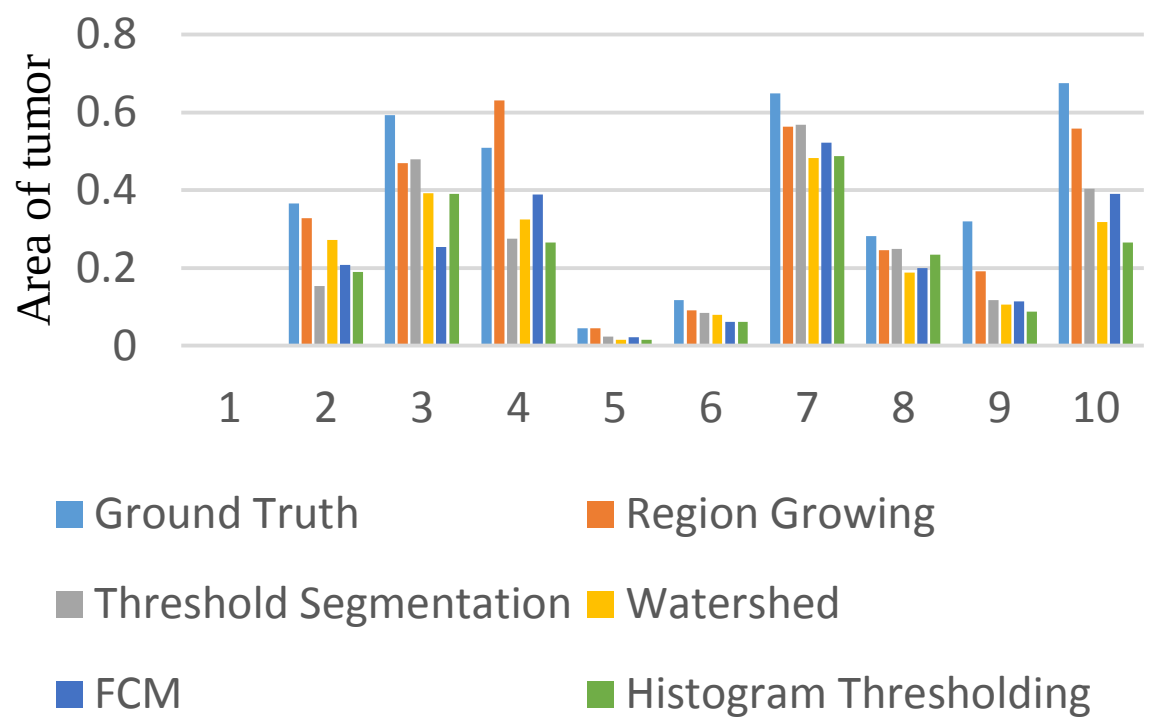

Figure 4: Comparison of Area of Tumor Segmented

Tumors are segmented out from images semi-automatically after performing threshold operation with different approaches. Figure.5. shows comparison made based on the average of area of tumor. As the area of tumor is different for different segmentation methods, however looking into Figure.5. Average of each case for each methods reveals that seeded region growing is performing better between all of above stated methods.

Because of its nature of grouping the pixels into regions, the infected regions are segmented out based on seed value that let the each region grows independently. This approach found effective but involve high computational complexity for brain tumor segmentation than other algorithms, especially in case of tissues and homogeneous regions [12]. In other methods based on thresholding, only intensity consideration and no relationship between pixels, extraneous pixels, or ignoring of solitary pixels, could loss information.

Wrong detection of tumor could be explored by small experiments shown in figure.7., for example each method have some kind of threshold value if the right threshold is not selected then probably the detection result would be wrong. For seeded region growing selection of seed point is the most important task, if the right seed is not allocated then result would be wrong.

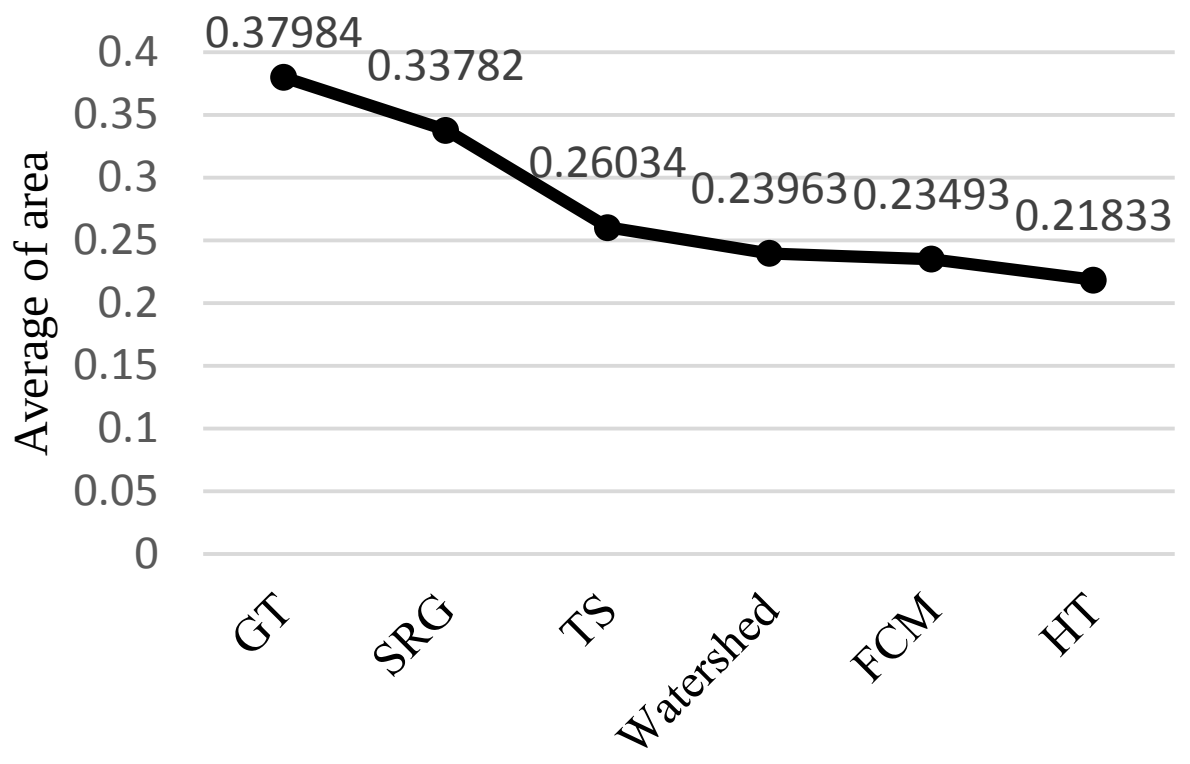

Figure 5: Comparison of Average of Area of Tumor

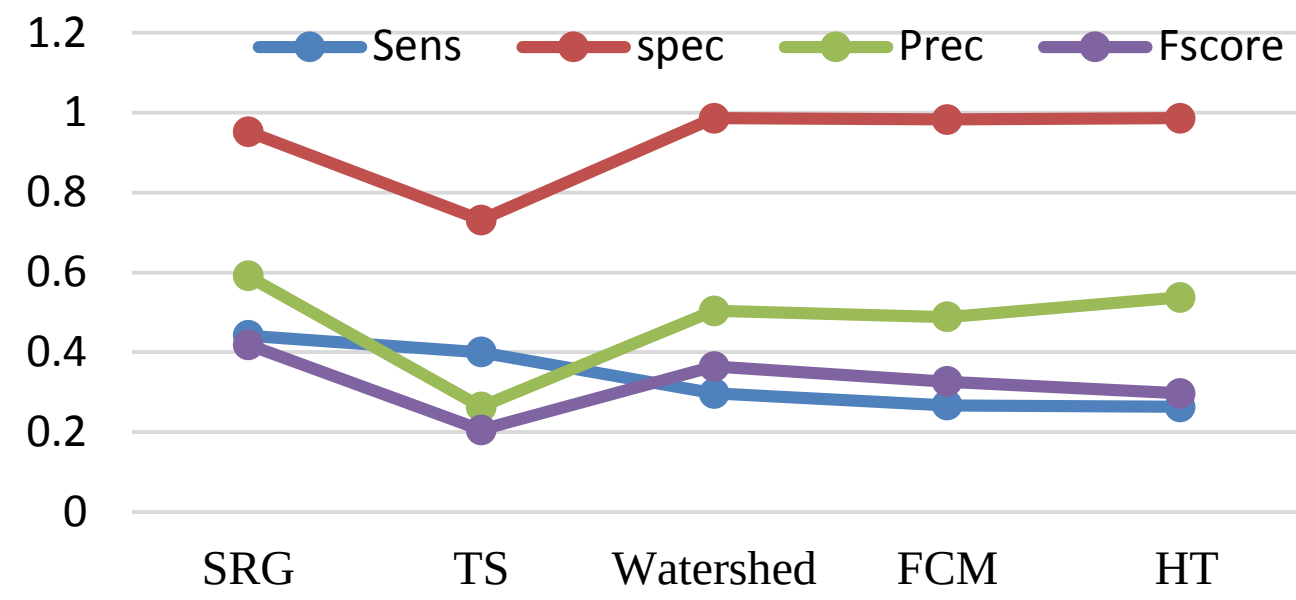

Figure 6: Comparison of Statistical Measures

Similarly for other four methods wrong threshold brought up with wrong results. Above statement is proved experimentally in figure.7. There are more than 120 types of tumors with different grades within same tumor. Every algorithm could not address each type of tumor, hence there is chance of wrong detection. In figure.8 visual comparison is shown

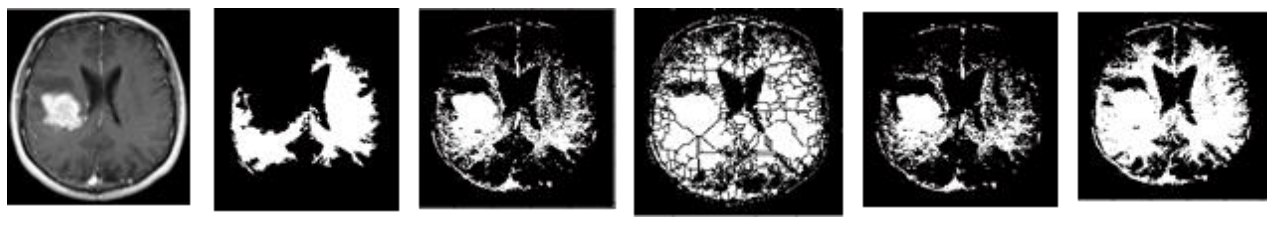

Figure 7: Wrong detection results at different threshold levels

## 4. Conclusions

Calculation of tumor area plays a vital role in assisting the treatment planning. Although the computer aided techniques are complex and require huge effort to be implemented but not as tedious, laborious, and time consuming as manual methods. In this paper five image segmentation methods have been implemented and compared to segment out the tumor from MR image dataset. Statistical and visual analysis proves that seeded region growing method is found to be best among all analysis methods. This method found best because of the nature of medical images and correctly segmentation of those regions having similar properties. In Future these algorithm could be tested for breast, lung, skin etc. We are planning to propose more robust and accurate segmentation algorithm for segmentation of tumor from brain, lung, etc. based on thresholding of images.

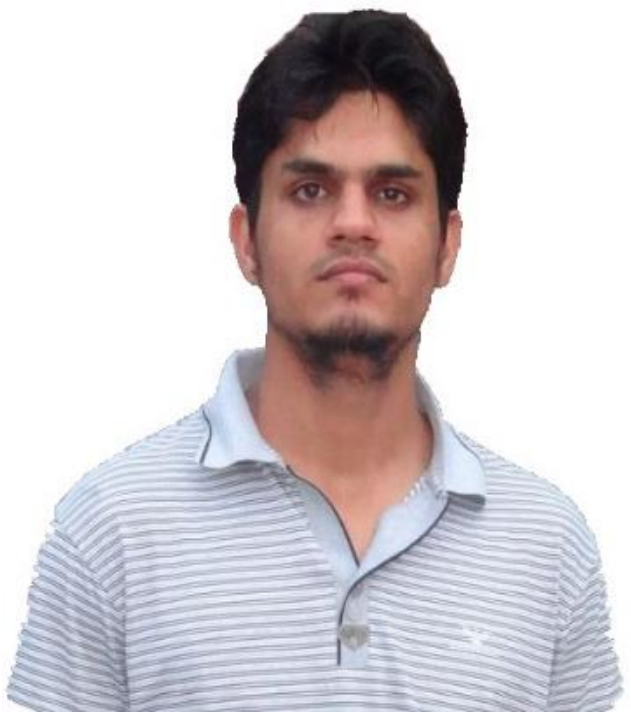

**Muhammad Ali Qadar**, born in 1987, Okara, Punjab, Pakistan. He earned BS in Computer Engineering in Sep 2011 from Bahaudding Zakariya, Multan Pakistan. He was serving ZDAAS LLC from Sep 2011 to Sep 2012. In Sep 2012 He was awarded Chinese Govt. Scholarship (CSC) and admitted to Beihang University in School of Electronic information Engineering. Based on his academic background and potential in research he awarded from Beijing Municipal Government at Beihang University, "Distinguished foreign student award 2013". He worked in Laboratory of Radar and Signal Processing at School of Electronic Information Engg., Beihang University, Beijing, China for almost two years from Sep 2012 to Oct 2014. His research interests are Image Segmentation, Enhancement and reconstruction in area of biomedical image processing. Currently, He is PHD student at Chinese Academy of Science, Institute of computing Technology, Beijing, China PR.

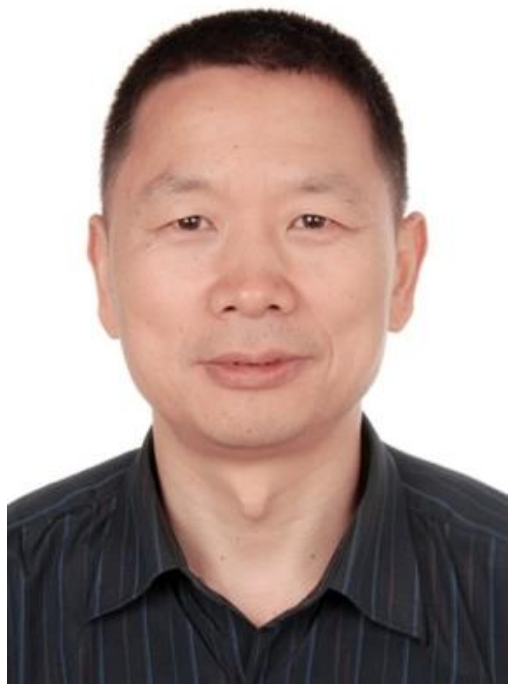

Zhaowen Yan received the B.Sc degree, the M.S. degree and the Ph.D degree in electrical engineering from Henan Polytechnic University, Xi'an University of Science and Technology and Xi'an Jiaotong University, China, in 1991, 1996 and 1999, respectively. From 1999 to 2002 he was a Post-Doctoral Research Associate at the Huazhong University of Science and Technology, where he conducted research on the development of the New-Style Equivalent Source Method of electromagnetic field, and the numerical simulation of the 3-D electromagnetic fields of aluminum reduction cells. From April 2003 to May 2005, he hold a Post-Doctorial position at Beihang University and from December 2012 to December 2013, he is a visiting scholar in EMC Laboratory, Missouri University of Science and Technology, U.S.A. He is now a professor of electronic science and technology at Beihang University. For many years, his research activity was focused on the electromagnetic field computation and EMC analysis. He has published over 70 conferences and journal papers and 7 monographs on electromagnetic theory. He has 17 authorized patents. He was awarded with a gold medal from Ministry of Education of PRC in 2003 and a gold medal from Ministry of Industry and Information Technology of PRC in 2011 as recognition of his research work. And the project he held won the second prize for the National Technological Invention in 2012.

## Appendix:

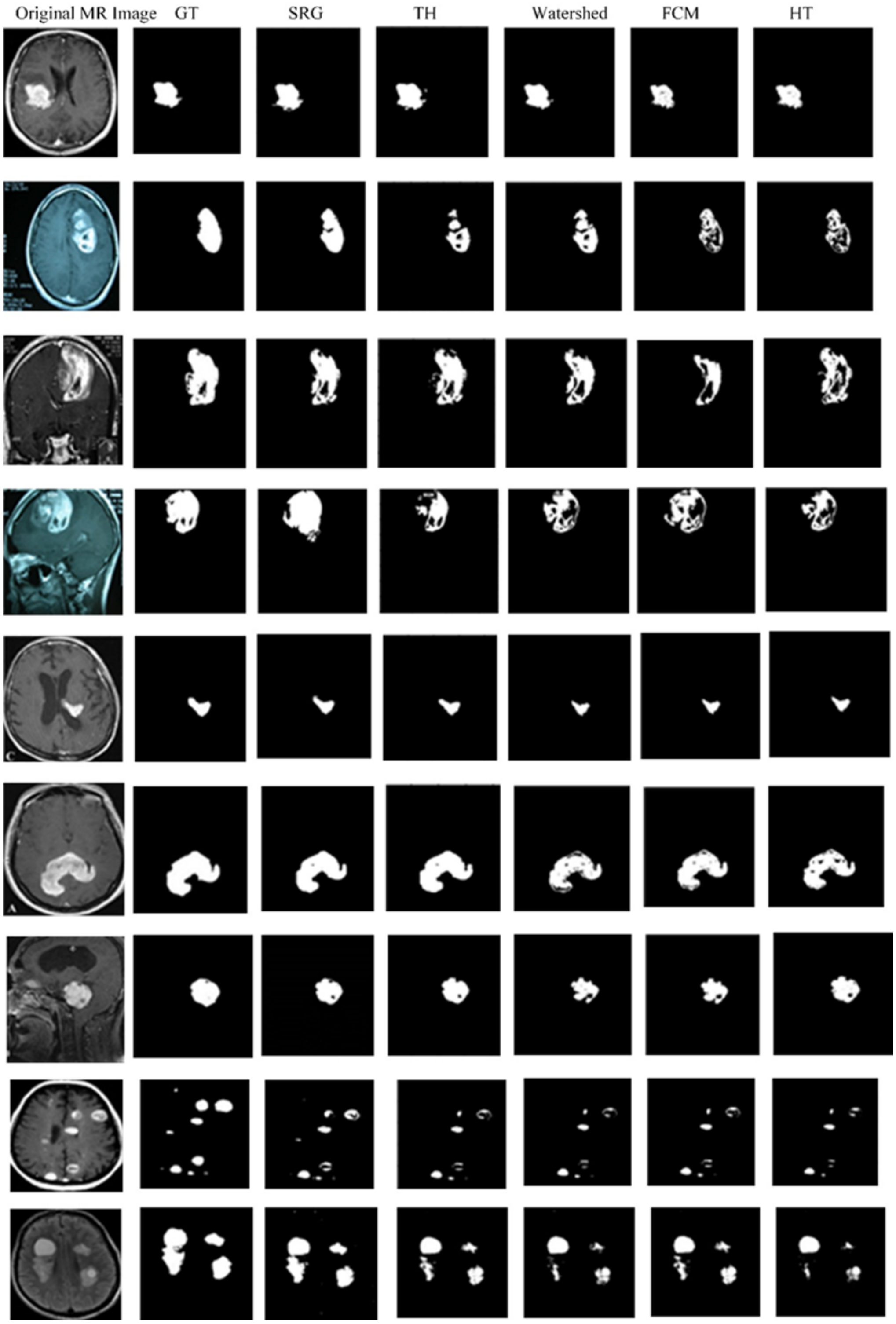

Figure 8: Visual Comparison of Performance